\documentclass[12pt, journal, letterpaper]{IEEEtran}
\usepackage{graphicx,subfigure,comment,tabularx}
\usepackage[table]{xcolor}
\usepackage{cite}
\usepackage{soul}
\usepackage{balance}

\newcommand{\Fig}[1]{Fig.~\ref{fig:#1}}
\newcommand{\Tab}[1]{Tab.~\ref{tab:#1}}
\newcommand{\Sec}[1]{Sec.~\ref{sec:#1}}
\newcommand{\ourthing}{FlexRel}

\begin{document}
	
\title{
Combining Relevance and Magnitude \\for Resource-saving DNN Pruning
}
\author{\IEEEauthorblockN{C.~F.~Chiasserini\IEEEauthorrefmark{1}\IEEEauthorrefmark{2}\IEEEauthorrefmark{3}\IEEEauthorrefmark{4}, F.~Malandrino\IEEEauthorrefmark{2}\IEEEauthorrefmark{3}, N.~Molner\IEEEauthorrefmark{5}, Z.~Zhao\IEEEauthorrefmark{1}\\
\small{
\IEEEauthorblockA{\IEEEauthorrefmark{1} Politecnico~di~Torino,~Italy --
\IEEEauthorrefmark{2} CNR-IEIIT,~Italy --
\IEEEauthorrefmark{3} CNIT,~Italy\\
\IEEEauthorrefmark{4} Chalmers~University of Technology,~Sweden --
\IEEEauthorrefmark{5} iTEAM Research Institute, Universitat Polit\`ecnica de Val\`encia, Spain
}
}
}
}

\maketitle

\begin{abstract}
Pruning neural networks, i.e., removing some of their parameters whilst retaining their accuracy, is one of the main ways to reduce the latency of a machine learning pipeline, especially in resource- and/or bandwidth-constrained scenarios. In this context, the pruning {\em technique}, i.e., how to choose the parameters to remove, is critical to the system performance. In this paper, we propose a novel pruning approach, called \ourthing\, and predicated upon combining training-time and inference-time information, namely, parameter magnitude and relevance, in order to improve the resulting accuracy whilst saving both computational resources and bandwidth. Our performance evaluation shows that \ourthing\, is able to achieve higher pruning factors, saving over 35\% bandwidth for typical accuracy targets.
\end{abstract}


\section{Introduction}
\label{sec:intro}

In the last years, computing at the edge has been gaining importance as more sectors get digitalized and require processing of data closer to the end users. This includes the storage of data as well as the intelligence, e.g., machine learning (ML) to process and extract knowledge from such information.
However, ML -- especially when implemented through deep neural networks (DNNs), has significant requirements. In addition to computational 
resource consumption in the data centers, critical issues are represented by the bandwidth consumption due to the information transfer on the radio link from the (potentially large number of) data sources and the training and inference time required by large  (ML) models for training and inference.  
Indeed, mobile  applications typically demand for a swift inference outcome, which can be challenging to obtain in a resource-constrained scenario like the network edge.  
The magnitude of the challenge is such that the SA6 group of 3GPP is discussing model transferring approaches to alleviate it.

ML model compression~\cite{hedge2023network, noi-infocom23, xie2020lite} has recently emerged as a way to address this crucial issue and a significant amount of work has leveraged this approach to find the optimal balance between the ML model size (hence, complexity) and performance. For instance,~\cite{noi-infocom23} developed an algorithm to compress ML models on-demand offering flexibility to run them on different devices at the cost of increasing time complexity. \cite{hinton2015distilling}~introduced the concept of knowledge distillation (KD) to compress the knowledge of a set of full and specialized models into a single one, improving accuracy in neural networks that are not very deep. \cite{jian2023communication}~considers pruning the DNN parts impacting most on the latency, ensuring shorter epoch training time, however more epochs may be required, incurring in similar global convergence time. \cite{zhang2018lq} investigates quantization based methods to obtain highly compact DNNs that can be efficiently computed at the price of decreasing the accuracy compared to  full-precision methods.

Among the above techniques, pruning represents a very appealing option due to its low complexity and, as noted above, it allows for small ML models while preserving high-quality inference as far as the adopted pruning (i.e., compression) factor
is not too high, and
no significant drift is observed in the input data statistics. The justification behind pruning is the so-called lottery-ticket hypothesis, introduced in the seminal paper~\cite{frankle2018lottery}: only a small portion of the DNN parameters have a significant impact on its effectiveness, hence, the rest can be safely removed. The hypothesis, however, does not say {\em which} are the ``winning tickets'', hence, several pruning {\em techniques} have been devised to identify -- and avoid removing -- them.

\begin{figure}
\centering
\includegraphics[width=\columnwidth]{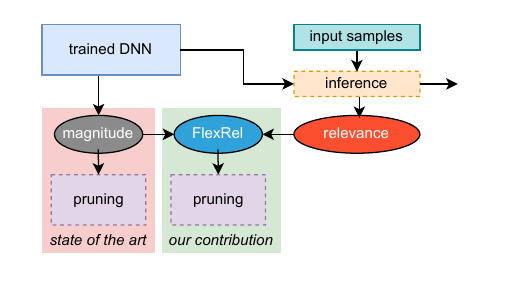}
\caption{
Three ways to make pruning decisions: the traditional way (on the left-hand side), i.e., directly considering the {\em magnitude} of the DNN parameters and pruning those with the smallest magnitude;   {\em relevance-based} pruning (on the right-hand side) -- with relevance being a quantity computed during inference --  using both the DNN parameters and input samples. Our \ourthing\ approach (in the middle), which  combines {\em both} magnitude and relevance to make more effective pruning decisions.
    \label{fig:flowchart}
} 
\end{figure}

\Fig{flowchart} summarizes the most relevant ways to identify the parameters to prune: filled boxes represent input information; patterned boxes correspond to operations; ellipses represent ways to combine or use output data.
The most common pruning technique is magnitude-based pruning~\cite{frankle2018lottery},
represented on the left-hand side of \Fig{flowchart}; it is fairly straightforward and simply removes
the lowest-magnitude ML model parameters.
In spite of its simplicity, magnitude-based pruning has been consistently shown to be fairly effective, and had been further refined through
alternative approaches that trade higher complexity for better accuracy.
Examples include works that use the {\em gradients} of parameters instead of their magnitude~\cite{shen2022prune}, and approaches that seek for an optimal {\em pruning threshold}, and remove all parameters whose magnitude falls below it. Other works instead solve an optimization problem to decide whether or not each parameter shall be pruned. 
Recent works extend pruning to emerging types of network, including recurrent ones and transformer architectures~\cite{kwon2022fast}. Finally, works like~\cite{hohman2024model} collect real-world experiments and experience with model pruning, and draw conclusions on the best strategies to maximize its effectiveness.

Additionally, there exist other techniques that, although not originally designed with pruning in mind, can be successfully adapted to it. A prominent example is parameter relevance, a methodology designed for Artificial Intelligence (AI) explainability that can be extended, as better explained in \Sec{sub-scores}, to DNN pruning.
Prominent among such techniques is {\em parameter relevance}. The relevance of each parameter expresses how much influence the value of each DNN parameter has on its final output; importantly, it also means that changing the value of a low-relevance parameter -- or, crucially, removing it altogether -- will not significantly change the DNN output. Thanks to this property, 
relevance scores can be used as an alternative to magnitude to select the parameters to prune, or both magnitude and relevance -- as opposed to magnitude alone -- can be considered when making pruning decisions.
As depicted on the right-hand side of \Fig{flowchart},
relevance scores and magnitude are obtained at different moments in time. Specifically, magnitude can be observed {\em a priori}, during the training of the DNN or immediately after; relevance, on the other hand, is computed {\em a posteriori}, considering a trained DNN and a specific set of input data. It follows that computing relevance could be more onerous than observing magnitude; on the positive side, relevance accounts for more valuable information.

Our main contributions, also highlighted in the center of \Fig{flowchart}, can be summarized as follows:
\begin{itemize}
    \item We identify {\em relevance} as a promising metric to exploit when making pruning decisions;
    \item We make the key observation that both high-magnitude and high-relevance parameters contribute to the learning performance, albeit through different mechanisms;
    \item We propose a new pruning approach, called \ourthing, exploiting both magnitude and relevance to select the parameters to prune;
    \item We validate our intuition through a set of experiments using publicly-available, popular DNNs and datasets, finding that \ourthing\ leads to a substantial performance improvement in terms of latency and bandwidth consumption at the cost of a modest amount of additional computing complexity;
    \item We further discuss when, given the features of the scenario at hand, it is beneficial to use \ourthing.
\end{itemize}

The rest of the paper is organized as follows. \Sec{current} discusses some relevant related work. \Sec{flexrel} explains our concept and methodology, while \Sec{results} validates our approach and demonstrates its superiority with respect to state-of-the-art alternatives. \Sec{open} highlights open issues for future research, and \Sec{conclusion} concludes the paper.

\section{Current Pruning Approaches}
\label{sec:current}

Pruning is a popular solution for ML model compression~\cite{zhong2022sensitivity, zhang2021unified}. The ambition of model pruning is to remove some model parameters either at the end of the training to get a smaller, less complex model, or after each epoch as the training proceeds (a.k.a. dynamic pruning) to also make each epoch duration shorter.
We remark that, as pruning reduces the size of an ML model, in the case of distributed training or inference, it can also significantly decrease the amount of information that needs to be transferred over the communication links connecting the nodes that contribute to the task as well as the overall latency.  Such scenarios have gained prominence recently, as ML models are deployed towards the edge of the network due to the fact that large amounts of data needed for training are collected at the edge  and  that data input to inference is  often generated at the end devices and ML-based applications require low latency. Further, even if an ML model is trained or fine-tuned in a cloud server, it may then need to be delivered to edge nodes, in which case, transmitting the model would imply  a non-negligible network load.

Notably, most
works on pruning focus on
pruning by magnitude, as it is a simple, yet effective, technique.
Essentially, if the magnitude of a given parameter is small, then its influence on the DNN output is limited, hence, it is preferable to prune that parameter rather than other, higher-magnitude ones. Key to this technique is the fact that weights can be cut {\em a priori}, i.e., just by observing the magnitudes without the need to compute any extra value.
Other works, however, propose new metrics for pruning, focusing on, e.g., the evolution of gradients~\cite{shen2022prune}. Focusing on distributed scenarios, \cite{jian2023communication}~accounts for communication issues when making pruning decisions, e.g., pruning away those parts of the DNN with the largest latency footprint.   

Alternative model compression approaches pursue the same goal by focusing on information compression~\cite{hedge2023network, noi-infocom23, xie2020lite}. In particular,  \cite{hedge2023network} proposes lossy {\em network compression} by representing with fewer bits the least important parameters of the DNN,
i.e., those deemed less likely to be winning tickets~\cite{frankle2018lottery}. Finally, \cite{xie2020lite} introduces semantic compression to preserve model explainability.

To the best of our knowledge,
{\bf existing pruning techniques only account for information available at {\em training} time}, e.g., the magnitude of DNN parameters and their evolution over training epochs. Furthermore, 
{\bf there is no work combining different  pruning techniques}, i.e., different ways of choosing the parameters to prune, to  increase the learning performance.

\section{The \ourthing\  Approach}
\label{sec:flexrel}

All pruning techniques are predicated on keeping the most important parameters in the DNN (i.e., those deemed most likely to be ``winning tickets'') and removing the rest.  As  depicted in \Fig{flowchart}, pruning techniques essentially differ in what quantities (e.g., parameter magnitude or gradients) are used to select the parameters to prune. Further, current pruning techniques exploit and combine information generated at {\em training} time, such as parameter magnitude; however, there is no need for this to be the case.

This is especially important in {\em split learning} scenarios, where mobile nodes and edge-based servers cooperate in running the same learning task, e.g., the training of a DNN. The mobile node hosts the local data and runs the first layers of the DNN, then, intermediate results are sent to the edge where the rest of the layers are hosted -- hence, the ``split'' name. In split learning scenarios, the total learning time (i.e., the duration of each epoch) is given by {\em two} components, to wit, the computation time and the network delay. High-quality pruning decisions might require more computation, but this can be compensated by the higher pruning fractions they allow, hence, the need to send less data over the network.

In this context, the
key intuition behind our \ourthing\ scheme is to exploit additional information, generated during the {\em inference} phase, to make pruning decisions. The reason is twofold:
\begin{enumerate}
    \item In general, considering both the training and inference phases allows gathering more information, hence, potentially making better decisions;
    \item Especially in scenarios where training and testing datasets can be qualitatively different~\cite{hedge2023network,noi-infocom23,jian2023communication}, the parameters that matter the most during inference (the so-called ``winning tickets'') may not be the same as those that evolve the most during training.
\end{enumerate} 
To leverage on our intuition, we need a way to quantify the importance of DNN parameters during inference; however, no ready-made metric akin to magnitude exists. Therefore, we first define a new metric based on relevance (\Sec{sub-scores}), and then discuss how to use it in \Sec{sub-using}.

\subsection{Inference-time importance of DNN parameters: Relevance scores}
\label{sec:sub-scores}

We begin by considering an aspect that is different from pruning but related to it, namely, {\em input relevance}. The basic goal of input relevance, which has been introduced in~\cite{montavon2018methods}, is to assess which parts of each input sample (e.g., which pixels of an image) had the largest impact on the DNN decision (e.g., the class assigned to the image). As an example, in the case of image classification, background pixels will have lower relevance than those belonging to an object. Input relevance is computed through the {\em activity maximization} framework, a set of mathematical techniques seeking for the input data (e.g., the pixels) that maximize a certain part of the output (e.g., the logit associated with the selected class). These techniques are applied recursively from the output layer of the DNN to its input.
The result of the operation is a {\em score} associated with each parameter of the DNN, expressing its impact on the output; it follows that scores also express how much the output itself would change if the parameter were removed. Importantly, such high-relevance parameters are not necessarily the highest-magnitude ones.

Input relevance is computed at inference time and accounts for inference-time input information, hence, it fits very well the goal of our \ourthing\ scheme to account for the inference phase as well. However, given the relevance scores defined for input information,  we need to understand how to exploit such scores to identify the DNN parameters to prune, i.e., we need to extend the notion of relevance to the model parameters. The extension is straightforward for fully-connected DNN layers where, denoting with~$n$ the size of the input vector, each element $b_{ij}$ of the output is given by 
$b_{ij}=\sum_{k=1}^n a_{ij}w_{kj}$,
i.e., a summation of products between elements~$a_{ij}$ of the input and model parameters~$w_{kj}$. 

We can interpret the above formula as {\em linking} parameters~$w_{kj}$ with elements~$a_{ij}$ of the input and elements~$b_{ij}$ of the output. Recalling that we can compute the relevance of both the input and the output as per~\cite{montavon2018methods}, we make the relevance of each parameter proportional to the relevance values of input and output elements linked to it.
Specifically, indicating with~$\mathsf{rel}$ the relevance, we have:
\begin{equation}
\label{eq:rel}
\mathsf{rel}(w_{kj})=\sum_{i=1}^n\left[\mathsf{rel}(a_{ik})+\mathsf{rel}(b_{ij})\right].
\end{equation}
Intuitively, if a parameter connects relevant inputs with relevant outputs, then it must be relevant itself.

It is worth remarking that, virtually, all modern DNNs consist of  types of layers other than fully-connected; most relevantly, image classification DNNs heavily feature convolutional layers. Notably, convolutional layers have been shown to admit equivalent fully-connected representations, hence, we can further extend our notion of parameter relevance to convolutional networks. 
For convolutional neural layers, we exploit the fact that, as shown in~\cite{ding2021repmlp}, convolution operations can be transformed into matrix products, hence, convolutional layers can be transformed into fully-connected ones (those used in multi-layer perceptrons). We can then use (\ref{eq:rel}) to compute the relevance of the transformed parameters.

\subsection{Using Relevance Scores}
\label{sec:sub-using}

Given their ability to capture the inference-time behavior of the DNN and to account for inference-time inputs,
one might be tempted to  take relevance as the sole criterion to select which parameters to prune, thus  altogether replacing magnitude with relevance.
However, in doing so, one might risk losing the information obtained during training and carried by magnitude values. Indeed, as we mentioned earlier, high-relevance parameters tend to have high-relevance elements (e.g., pixels) as {\em both} input and output. At the same time, high-magnitude parameters might connect high- and low-relevance ones elements, precisely because of their large magnitude. Importantly, {\em both} actions are important parts of the way DNN operate, and both need to be preserved when performing pruning.

Indeed, relevance and magnitude seek to ask the same question, i.e., which parameters of the DNN are the most important, albeit in different ways. As reported in \Tab{features}, \ourthing\ seeks to combine both into a single score, aiming at keeping only those parametes that exhibit 
both high relevance and high magnitude parameters. Specifically, for each parameter,
magnitude and relevance
are first normalized between~$0$ and~$1$, and then summed in a weighted manner,
with~$\delta$ indicating the weighting factor.
The weighting factor~$\delta$ also allows for managing conflicts or inconsistencies between relevance and magnitude. Specifically:
\begin{itemize}
    \item Parameters with high magnitude and high relevance will have very high score, close to~$1$ (the maximum possible);
    \item Parameters with low magnitude and low relevance will have very low score, close to~$0$ (the minimum possible);
    \item Parameters with low magnitude and high relevance will have a score close to~$\delta$;
    \item Parameters with high magnitude and low relevance will have a score close to~$1-\delta$.
\end{itemize}
It follows that, by selecting~$\delta$, we can also decide how high (or low) the score of parameters where relevance and magnitude do not match will be.

The main features of the pruning techniques we consider are summarized in \Tab{features}.
It is worth mentioning that traditional, magnitude-based pruning already works reasonably well in many cases, thanks to the extensive research devoted to it, as reviewed in \Sec{current}. Accordingly, our goal with \ourthing\ is not to replace existing methodologies, but rather to complement and perfect them.

\begin{table}[h!]
\caption{
Comparison between pruning techniques
\label{tab:features}
} 
\begin{tabularx}{1\columnwidth}{|c||X|X|X|}
\hline
Strategy & SoTA: & Benchmark: & Ours:\\
feature & magnitude & relevance & \ourthing
\\\hline\hline
inputs & magnitude~$M$ & relevance~$R$ & both~$M$ and~$R$
\\\hline
score~$s$ & $s{\gets} M$ & $s{\gets} R$ & $s{\gets} \delta M$ ${+}$ $(1{-}\delta)R$
\\\hline
requires relevance & \cellcolor{green!25}no & \cellcolor{orange!25}yes & \cellcolor{orange!25}yes
\\\hline
effectiveness & \cellcolor{blue!25}good & \cellcolor{orange!25}medium &  \cellcolor{green!25}best
\\\hline
best suited for & CPU-constrained scenarios & & balanced scenarios
\\\hline
\end{tabularx}
\end{table}

In the next section, we explore the impact of the weighting factor $\delta$ -- expressing the relative importance we assign to relevance and magnitude.

\section{Numerical Results}
\label{sec:results}

In the interest of reproducibility, we design our reference scenario using free, popular, and publicly-available models, datasets, and information. Specifically, our experiments use the VGG16 DNN and the ImageNet dataset for image classification. VGG16 is a convolutional DNNs widely used for computer vision applications, with 16~layers and about 138~million parameters. ImageNet is a popular dataset including 1,000~classes and over 1.2~million images, and is one of the most challenging -- and most used -- benchmarks when performing image classifications. We use the Python programming language, the PyTorch framework, and the Adam optimizer, with a starting learning rate of~$0.05$.

We perform a total of 50~training epochs: 10~with the full DNN (i.e., before pruning), and 40~with the pruned DNN (i.e., after pruning). All information used for pruning (magnitude and/or relevance) is computed after the first 10~epochs.
Pruning is always structured, e.g., pruning decisions concern entire channels (``filters'') of convolutional layers and not individual parameters.
We compare the following three pruning {\em techniques}, i.e., ways to choose the filters to remove, exploiting different kinds of information:
\begin{itemize}
    \item[{\em (i)}] {\em Magnitude}, i.e., the SotA approach of removing the filters with the lowest average magnitude;
    \item[{\em (ii)}]  {\em Relevance}, i.e., removing the filters with the lowest relevance;
    \item[{\em (iii)}]  Our {\em \ourthing} approach, combining both relevance and magnitude as previously explained.
\end{itemize}
The three techniques correspond, respectively: {\em (i)} to the state of the art approach~\cite{jian2023communication,frankle2018lottery,zhang2021unified}, {\em (ii)} to an approach based exclusively on the notion that relevance can be a good guidance in pruning decisions, and {\em (iii)} an approach integrating such a notion within existing, well-established, well-performing pruning methodologies.

The connectivity between the mobile device and edge node is modeled accounting for typical 5G data rates\footnote{Source: OpenSignal.com, ``Benchmarking the Global 5G Experience -- June 2023''.} of 150~Mbit/s in uplink and 20~Mbit/s in downlink. As in all split learning scenarios, the mobile device and the edge node have to exchange the input/output of the so-called cut-layer at each epoch.

\begin{figure}[h!]
\centering
\includegraphics[width=.8\columnwidth]{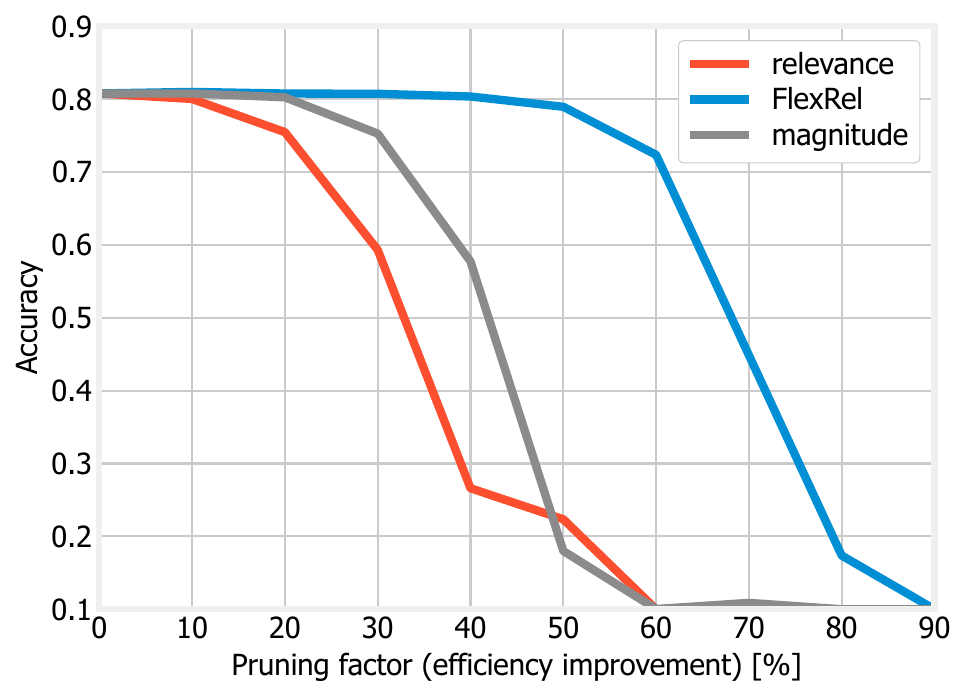}
\caption{
Accuracy reached by the VGG16 DNN when trained over the ImageNet dataset as a function of the pruning factor, for different pruning techniques.
    \label{fig:accuracy}
} 
\end{figure}

{\bf Roadmap.}
We study the impact of three main quantities: {\em  (i)} the pruning factor, {\em (ii)} the accuracy target, and {\em 
 (iii)} the weight factor~$\delta$. Specifically, through our performance evaluation, we characterize the relationship between each of these quantities and the overall behavior of the system, and  how pruning strategies influence such relationships.

\Fig{accuracy} reports the accuracy achieved as a function of the pruning factor, for different pruning techniques.
Recall that pruning a certain fraction of the weights reduces, by virtually the same fraction, both the computations to perform and the quantity of data to transfer across nodes. Pruning fractions thus map, one-to-one, to efficiency improvements. We can distinguish three main regions in each of the curve:
\begin{itemize}
    \item At first, for small pruning factors, the accuracy remains almost constant;
    \item As pruning factors grow, the accuracy drops in a roughly linear manner;
    \item Further increasing pruning, the accuracy stays constant to a very small value.
\end{itemize}
The third region shall always be avoided; in most cases, we want to operate around the border between the first and second region.

We can observe that using magnitude alone provides  very good performance, which is consistent with the popularity of this  approach. Using relevance alone results, instead, in a substantially lower accuracy. Intuitively, only considering that metric results in pruning
high-magnitude parameters, which adversely affects learning performance.
Most importantly, {\em combining} relevance and magnitude as per our \ourthing\ approach results in the best performance, even better than magnitude. This key  result validates our intuition: magnitude and relevance express two different -- albeit related  -- quantities, {\em both} of which have a bearing on the final learning quality. Consequently, considering both when making pruning decisions results in the best performance. As an example, considering an accuracy target of~70\%, we are able to prune around 25\%~of parameters if we consider relevance, around 35\%~of parameters if we consider magnitude and over~60\% with \ourthing, for an overhead reduction of over~30\%.

\begin{figure}[h!]
\centering
\subfigure[\label{fig:times-sota}]{
\includegraphics[width=.8\columnwidth]{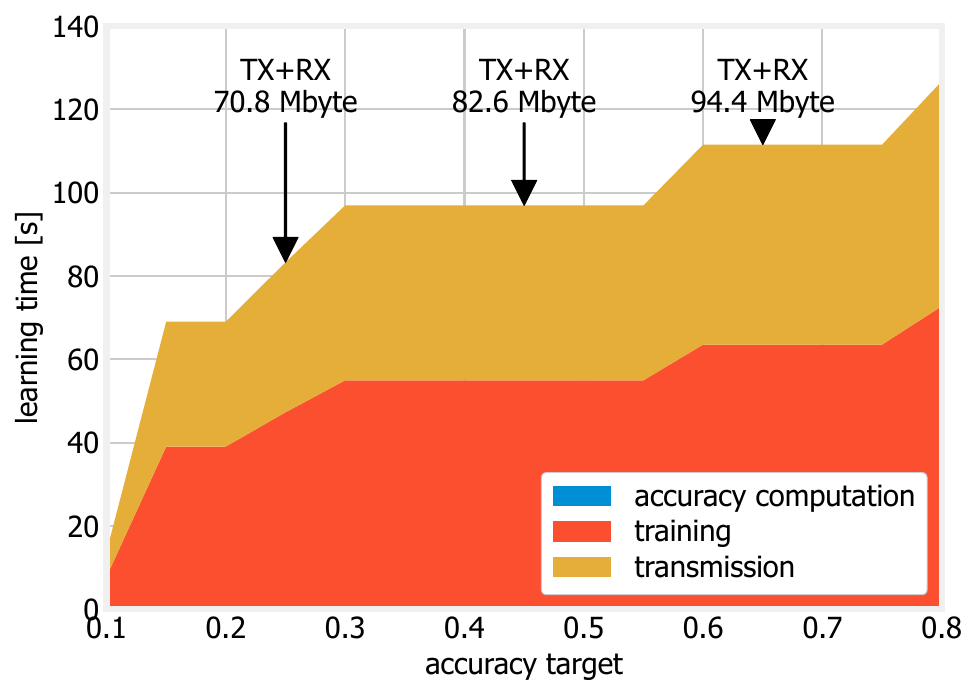}
} 
\subfigure[\label{fig:times-us}]{
\includegraphics[width=.8\columnwidth]{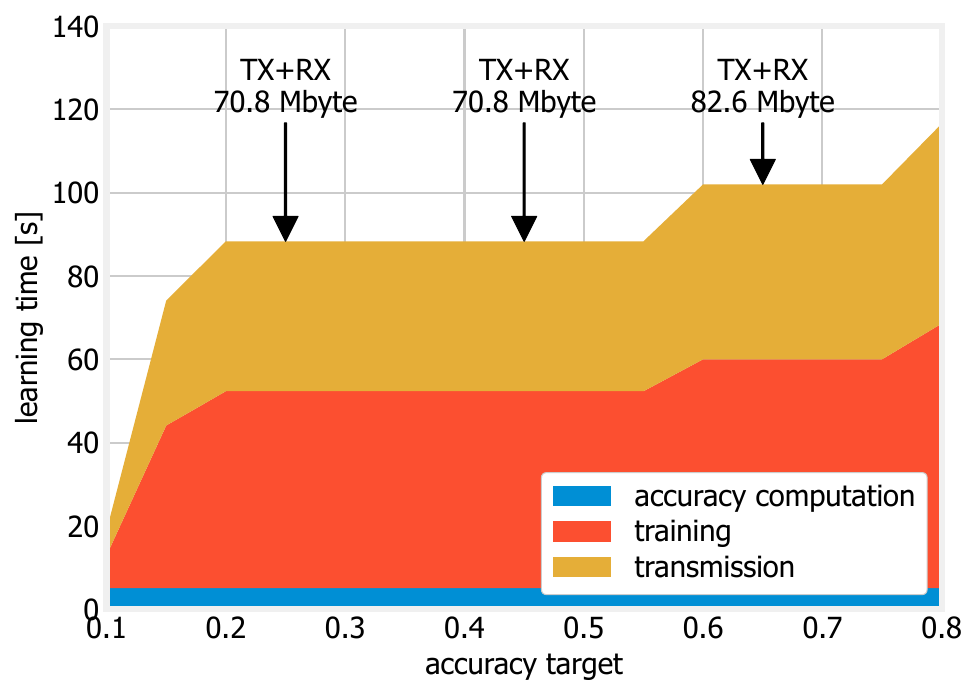}
} 
\caption{
Elapsed learning time as a function of the accuracy target, for the magnitude-based (a) and \ourthing\ (b) techniques. Numbers in the plot represent the quantity of transmitted data.
    \label{fig:times}
} 
\end{figure}

\Fig{accuracy} also shows that \ourthing\ pushes to the right the border between the first and the second of the three regions outlined earlier. In other words, it possible to attain higher pruning factors (hence, higher efficiency), exceeding 30\% and up to 50\%, without significantly degrading the learning performance (i.e., accuracy).
On the negative side, computing the relevance requires additional time, which might in principle negate the accuracy gains highlighted in \Fig{accuracy}. To verify whether or not this is the case, we vary the {\em accuracy target} between~10\% and~80\%, and quantify, for each technique, (i) how long it takes to reach that accuracy level, and (ii) how such time is spent. The results for the magnitude-based and \ourthing\ approach are summarized in \Fig{times} (the relevance-based technique is not represented due to its lower performance).

Next, we look at training times and investigate to which extent the additional computations needed to obtain relevance values impact the total elapsed time. \Fig{times} shows how the learning time changes as a function of the accuracy target, for the magnitude-based and \ourthing\ techniques.
We can indeed observe an additional contribution coming from the need to compute the relevance (blue area at the bottom). However, such a contribution is almost negligible compared to the time it takes to perform the actual training, i.e., the forward- and backward-passes (red areas) and the network delays (yellow areas). Comparing the total height of the colored areas between the two plots, we can see that \ourthing\ (\Fig{times-us}) results in substantially shorter learning times than  the magnitude-based technique (\Fig{times-sota}), in spite of the additional complexity.

The reason can be again inferred from \Fig{accuracy}: given a target accuracy level, e.g., 0.5, \ourthing\ reaches that level with a much higher pruning factor (in the example, 70\%) than magnitude-based pruning (in the example, 45\%). More aggressive pruning results in both less computation being performed (hence, shorter computation times) and less data being transferred between learning nodes and learning server (hence, shorter network delays). These two factors abundantly offset the extra time required to compute the relevance.

Focusing on the quantity of transmitted data, reported in \Fig{times}, we can observe that \ourthing\ results in a smaller quantity of transmitted data, which is consistent with the shorter time spent for network transmissions. Importantly, transmitting less data over the air also results in further benefits, including lower energy consumption, less congestion, and smaller likelihood of communication issues.

\begin{figure}[h!]
\centering
\includegraphics[width=.8\columnwidth]{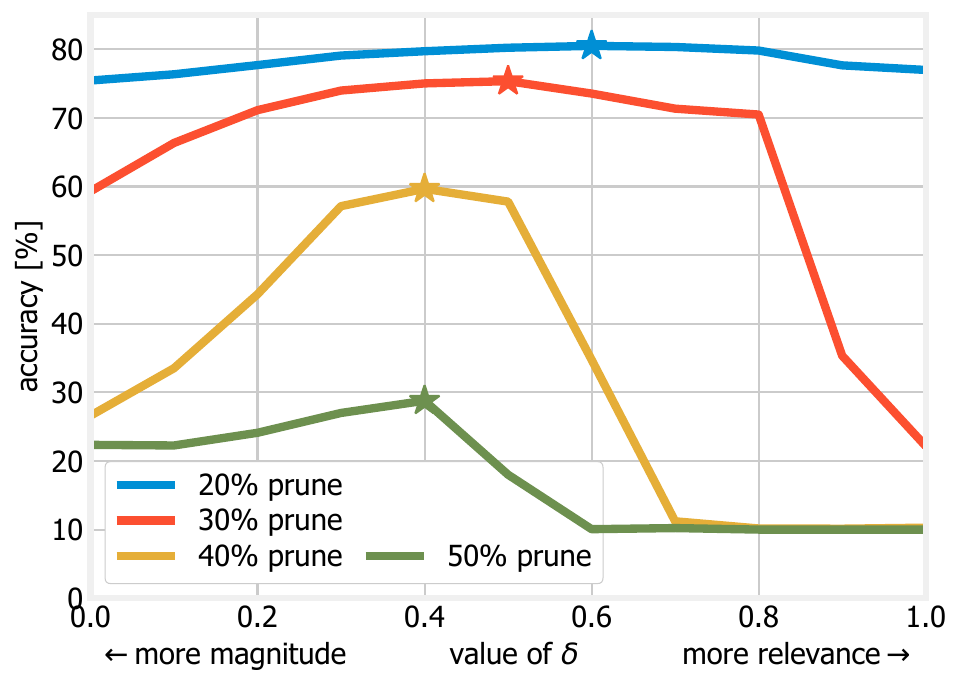}
\caption{
Effect of the weighting factor~$\delta$ over the achieved accuracy, for different pruning factors.
    \label{fig:deltas}
} 
\end{figure}

Last, in \Fig{deltas} we study the effect of the weighting factor~$\delta$ over the system performance. We can observe that the best performance is obtained for intermediate values of~$\delta$. In other words, both considering magnitude alone (i.e., $\delta{=}0$) or  relevance alone (i.e., $\delta{=}1$) results in suboptimal performance; in the latter case, the accuracy might become particularly low. Also notice how the best value of~$\delta$, identified by a star in the plots, depends upon the pruning factor, hence, some fine-tuning might be required.

In summary, we can observe that our \ourthing\ approach allows for (i) better learning quality given a pruning factor, and (ii) shorter learning times -- accounting for all contributions thereof -- given a learning quality target. It thus minimizes the potential drawbacks of pruning (i.e., learning quality degradation) while amplifying its benefits (i.e., shorter learning times).

This is due essentially to \ourthing's ability to account for multiple factors, i.e., relevance and magnitude, when making pruning decisions. Indeed, relevance and magnitude contribute to the training quality in different ways. Specifically:
\begin{itemize}
    \item High-relevance parameters tend to have high-relevance elements (e.g., pixels) as {\em both} input and output;
    \item High-magnitude parameters can {\em change} high-relevance elements into low-relevance ones, and vice versa.
\end{itemize}
By accounting for both metrics, \ourthing\ can avoid removing parameters that do either thing, hence, minimize the impact of pruning itself on the global learning quality.

\section{Discussion and Open Issues}
\label{sec:open}

Our experiments have shown
the benefits of combining different
methodologies 
-- exploiting both training- and relevance-time information --
to quantify the importance of parameters to improve pruning performance, to wit, magnitude and relevance. By doing so, our \ourthing\ scheme can improve both learning quality and training time. In addition to being a viable, effective technique on its own right, \ourthing\ and our analysis thereof point out several avenues for further investigation, as discussed next.

An important research direction consists in  considering additional metrics, beyond magnitude, to choose the parameters to prune. Indeed, as discussed in \Sec{current}, some existing works have sought to replace magnitude by gradient evolution~\cite{shen2022prune} or even {\em ad hoc} scores.
Taken in isolation, such alternative metrics have been found to perform similar (or slightly better) than magnitude. Therefore, it would be interesting to assess whether combining those metrics with relevance brings the same benefit we have observed with \ourthing.

A second area worthy of investigation is the dynamic aspect of pruning. As mentioned above, computing parameter relevance is a computationally inexpensive task, hence, it could be done at multiple training epochs. This would make it possible to leverage the {\em evolution} of relevance to make pruning decisions, in a manner similar to what works like~\cite{shen2022prune} do with gradients. In addition to the potential performance benefit, such a study would shed further light on the evolution of parameter relevance across epochs, hitherto a poorly explored topic.

In scenarios where learning is performed in a distributed manner (e.g., through FL), the relevance of both parameters and inputs can be exploited as a way to assess the value of local datasets -- hence, the nodes owning them -- to the overall training. Intuitively, if the local parameters of a certain node and/or its local dataset have low relevance, then that node might not give a substantial contribution to the distributed training. This can be of great help in scenarios when {\em node selection}~\cite{malandrino2021toward} is a crucial task, owing to either cost or learning time considerations.

Finally, we remark that the advantages of computing relevance may be limited in two main cases. The first is represented by nodes with very constrained computational capabilities, which simply cannot afford to compute relevance (e.g., their battery would deplete if they tried). The second case  includes those scenarios where relevance {\em could} be computed, but that would not help because very little or no pruning could be performed anyway, e.g., because the number of parameters in the DNN model is already very small. With reference to \Fig{accuracy}, in those scenarios accuracy would drop very quickly as the pruning factor grows, hence, choosing more carefully the parameters to prune provides no additional gain.

In a general sense, all the research directions sketched above contribute towards a {\em holistic} view of ML, where the way each learning task is approached accounts for such elements as the available data, the node(s) performing the training, and the model to use.

\section{Conclusion}
\label{sec:conclusion}
This paper proposes a new pruning approach, \ourthing, combining the most widely used metric for pruning DNNs, i.e., {\em magnitude}, with a non-traditional pruning metric, {\em relevance}; in doing so, it is able to leverage both training- and inference-time information. Such a combination offers better performance than each of the metrics individually used, as magnitude and relevance express two different -- albeit related -- properties of DNN parameters, both of which have an impact on the final learning quality.

As shown by our  experimental results, both actions are important parts of the way DNNs operate, and \ourthing\ keeps both, allowing for (i) better learning quality given a pruning factor, and (ii) shorter learning times -- accounting all contributions thereof --, given a learning quality target with (iii) reduced latency and bandwidth consumption. It thus minimizes the potential drawbacks of pruning (i.e., affecting learning quality) while amplifying its benefits (i.e., shorter learning times).

\section*{Acknowledgments}
The work was partially supported by the Smart Networks and Services Joint Undertaking (SNS JU) under the European Union's Horizon Europe research and innovation programme under the MULTIX project Grant Agreement No.\,101192521.
The research leading to these results has been partially funded by the Italian Ministry of University and Research (MUR) under the PRIN 2022 PNRR framework (EU Contribution -- NextGenerationEU -- M. 4,C. 2, I. 1.1), SHIELDED project, ID P2022ZWS82.
This work has been partially supported by the Spanish Ministry of Economic Affairs and Digital Transformation and the European Union-NextGenerationEU through the UNICO 5G I+D ADVANCING-5G-TWINS (Grant Agreement No. TSI-063000-2021-112, TSI-063000-2021-113, TSI-063000-2021-114) projects.

\bibliographystyle{IEEEtran}
\bibliography{refs}%
\vskip -1cm plus -1fil
\begin{IEEEbiographynophoto}
{Carla~Fabiana Chiasserini}
(M'98, SM'09, F'18)  is currently a Full Professor  with  the  Department  of
Electronic  Engineering  and  Telecommunications at Politecnico di
Torino, a WASP Guest Professor at Chalmers University of Technology, and a CNIT and CNR Research Associate.   
Her research interests include edge computing, network support to machine learning, and design and performance
analysis of mobile  networks and services. 
\end{IEEEbiographynophoto}
\vskip -1cm plus -1fil
\begin{IEEEbiographynophoto}
{Francesco Malandrino}
(M'09, SM'19) earned his Ph.D. degree from Politecnico di Torino in 2012
and is now a researcher at the National Research Council of Italy
(CNR-IEIIT). His research interests include the architecture and
management of wireless, cellular, and vehicular networks.
\end{IEEEbiographynophoto}
\vskip -1cm plus -1fil
\begin{IEEEbiographynophoto}
{Nuria Molner} obtained her Ph.D. from Universidad Carlos III de Madrid in 2021. Currently, she is a researcher at iTEAM Research Institute of Universitat Polit\`ecnica de Val\`encia (iTEAM-UPV).
\end{IEEEbiographynophoto}
\vskip -1cm plus -1fil
\begin{IEEEbiographynophoto}
{Zhao Zhiqiang}
earned his master degree from Politecnico di Torino in 2024. He is currently a research fellow at Politecnico di Torino.
\end{IEEEbiographynophoto}
\end{document}